\documentclass[letterpaper, 10 pt, conference]{ieeeconf}  

\IEEEoverridecommandlockouts                              

\usepackage{amsmath,amsfonts}
\usepackage{algorithmic}
\usepackage{algorithm}
\usepackage{array}
\usepackage{booktabs}
\usepackage{subcaption}
\usepackage{textcomp}
\usepackage{stfloats}
\usepackage{url}
\usepackage{verbatim}
\usepackage{graphicx}
\usepackage{cite}
\usepackage{xcolor}
\usepackage{dsfont}
\usepackage{multirow}
\usepackage{float}

\title{\LARGE \bf
	Evaluating a GAN for enhancing camera simulation for robotics
}

\author{Asher Elmquist$^{1}$, Radu Serban$^{1}$ and Dan Negrut$^{1}$
	\thanks{$^{1}$all authors are with the Department of Mechanical Engineering, University of Wisconsin-Madison, WI, USA
		{\tt\small \{amelmquist,serban,negrut\}@wisc.edu}}%
}

\definecolor{arsenic}{rgb}{0.23, 0.27, 0.29}
\definecolor{charcoal}{rgb}{0.21, 0.27, 0.31}
\definecolor{hanblue}{rgb}{0.27, 0.42, 0.81}
\definecolor{blue-ncs}{rgb}{0.0, 0.53, 0.74}
\definecolor{awesome}{rgb}{1.0, 0.13,0.32}
\definecolor{darkgreen}{rgb}{0, .4,0}
\definecolor{purple}{rgb}{.55, .2,.9}

\newcolumntype{M}[1]{>{\centering\arraybackslash}m{#1}}

\newcommand{\SBELsimnet}{\textit{$Net_{sim}^{nominal}$}}
\newcommand{\SBELrealnet}{\textit{$Net_{real}$}}
\newcommand{\SBELepenet}{\textit{$Net_{sim}^{enhanced}$}}

\newcommand{\SBELcitynetpt}{\textit{$Net^{PT}_{city}$}}
\newcommand{\SBELcitynet}{\textit{$Net_{city}$}}
\newcommand{\SBELgtanet}{\textit{$Net_{GTA}$}}
\newcommand{\SBELgtaepenet}{\textit{$Net_{GTA-EPE}$}}

\newcommand*{\rom}[1]{\expandafter\@slowromancap\romannumeral #1@}

\begin{document}

\maketitle
\thispagestyle{empty}
\pagestyle{empty}

\begin{abstract}
Given the versatility of generative adversarial networks (GANs), we seek to understand the benefits gained from using an existing GAN to enhance simulated images and reduce the sim-to-real gap. We conduct an analysis in the context of simulating robot performance and image-based perception. Specifically, we quantify the GAN's ability to reduce the sim-to-real difference in image perception in robotics. Using semantic segmentation, we analyze the sim-to-real difference in training and testing, using nominal and enhanced simulation of a city environment. As a secondary application, we consider use of the GAN in enhancing an indoor environment. For this application, object detection is used to analyze the enhancement in training and testing. The results presented quantify the reduction in the sim-to-real gap when using the GAN, and illustrate the benefits of its use.
\end{abstract}

\section{Introduction}
\label{sec:intro}
Generative adversarial networks (GANs) have shown remarkable versatility when performing image-to-image style translation. While these are typically demonstrated on benchmark real image datasets, herein GANs are used to enhance simulated images for robotics simulation with an eye towards reducing the well-documented sim-to-real gap \cite{ros2016synthia,lehman2018surprising,muratore2018domain,muratore2019assessing,langford2019applying}. At a high level, we are interested in simulating a robot operating in a complex environment \cite{artatk2022}. A prerequisite for this is the ability to simulate a camera sensor that produces faithful simulated images as the virtual robot moves around in the virtual world.

There are three prominent characteristics in GAN results that are detrimental in camera simulation: temporal inconsistency, artifacts, and modification of the ground truth including introduction of phantom objects. Temporal consistency is important for closed-loop simulation in robotics since the response of the perception algorithm (e.g., tracking) can be interrupted by inconsistent simulation \cite{sunderhauf2018limits}. Secondly, artifacts in GANs can lead to poor outcomes due to the high sensitivity of many perception algorithms to input data. For instance, artifacts not present in the training domain of an object detection algorithm could produce undesired effects. In many GANs, these artifacts must be explicitly considered as they are small enough to be obfuscated by low-resolution comparison. Lastly, the ground truth of the image must not be modified by the GAN. This is because the images are automatically labeled in simulation, and changes to the content without modification of the label will lead to erroneous results. Moreover, a main benefit of simulation is the ability to create corner cases and specify particular scene setups. If the GAN is allowed to change the labels, this key aspect of simulation is undermined.

Recent GAN research has targeted the three aspects outlined above. In the contribution ``Enhancing Photorealism Enhancement'' \cite{richter2021enhancing}, the GAN (herein called EPE or EPE-GAN) showed significant improvement relative to other GANs in terms of the visual metrics Kernel Inception Distance (KID) \cite{binkowski2018demystifying} and semantically aligned Kernel VGG Distance (sKVD) \cite{richter2021enhancing}. These two measures along with other visual metrics including Inception Score (IS) \cite{salimans2016improved} and Fr\'echet Inception Distance (FID) \cite{heusel2017gans}, are often chosen to quantify GAN improvements, in part, due to the metrics' correlation with human perception \cite{binkowski2018demystifying,salimans2016improved,heusel2017gans,wang2018high}. Indeed, GAN improvements are often demonstrated directly through studies of human perception \cite{richter2021enhancing,salimans2016improved,park2019semantic}. However, in simulation for robotics we are instead interested in what a GAN can do for robot perception, as we seek to provide a simulation that can be used in robotics for training and assessing robotic algorithms. In this work we consider assessment and training of image-based perception.

There are two primary ways in which simulation is used in autonomy stack design: (i) assessment/testing of an existing algorithm (typically trained on real data); and (ii) development/training of a new algorithm. For (i), the sim-to-real gap is the discrepancy between the predicted performance in simulation and the actual performance in the target application. For (ii), the sim-to-real gap is a domain transfer gap associated with the trained performance in simulation and the performance in the target application. In both cases, the performance measure is equivalent -- the performance difference between the simulated application and the real application. 

In this contribution, we use the EPE-GAN in simulation and report sim-to-real differences for assessment and training, i.e., both (i) and (ii) above. For assessment, we quantify the simulation's ability to predict performance of a real-data trained algorithm {\SBELrealnet} in a real application. However, for training, we quantify the simulation's ability to predict the performance of a sim-trained algorithm {$Net_{sim}$} in a real application. In both cases, we are interested in the difference between the predictive power of nominal simulation and the predictive power of simulation when enhanced with EPE-GAN. The contributions of this work can be summarized as: 
\begin{enumerate}
	\item quantitative evaluation of EPE-GAN as a tool to close the sim-to-real gap in camera simulation
	\item generalization of the contextualized performance (CPerf) validation methodology \cite{elmquist2022performance} for semantic segmentation as a measure for the sim-to-real gap
	\item demonstration of EPE-GAN for reducing the sim-to-real gap for a specific application of simulation for robotics  
\end{enumerate}

\subsection{Related Work}
While the GAN considered in this paper is EPE-GAN, many other GANs exist for image-to-image conversion, image synthesis, and image stylization, including CyCADA~\cite{hoffman2018cycada}, MUNIT~\cite{huang2018munit}, TSIT~\cite{jiang2020tsit}, CUT~\cite{park2020contrastive}, CycleGAN~\cite{zhu2017unpaired}, Pix2Pix~\cite{isola2017image}, SPADE~\cite{park2019semantic}, PhotoWCT~\cite{li2018closed}, WCT2~\cite{yoo2019photorealistic}, Color transfer~\cite{reinhard2001color}, and CDT~\cite{pitie2007automated}. See \cite{richter2021enhancing} for a comparison among some of these techniques. Each of these algorithms has its strengths, but for the purposes of simulation, EPE-GAN demonstrated better consistency of synthesized data than other methods. 

For quantifying the realism of the data produced by these algorithms, IS, FID, and KID are the most widely used metrics~\cite{huang2018munit,jiang2020tsit,park2020contrastive,salimans2016improved,karras2019style,zhao2020fine,richter2021enhancing}. Recently, as part of EPE-GAN, sKVD showed additional beneficial characteristics~\cite{richter2021enhancing}. All of these metrics consider the difference in images viewed with pretrained networks for unrelated tasks, and are motivated in part by their correlation with human perception. For using enhanced simulation data for training and evaluation perception algorithms, \cite{soufi2019data} considered the use of a GAN to produce synthetic signs for classification. A GAN was also used to improve the realism of underwater images~\cite{sung2020realistic}. GANs were used to improve the learning of real-world tasks~\cite{ho2021retinagan,du2022bayesian}, but the focus was on producing algorithms with high success in reality rather than producing data similar to reality. This difference is related to the difference between realistic simulation and domain randomization/dataset augmentation. CyCADA was also compared using a segmentation task, but with relatively poor results using a more summary comparison~\cite{hoffman2018cycada}.

\section{GAN enhancement for simulating Cityscapes environment}
\label{sec:cityscapes_and_gta}
\begin{figure*}
	\centering
	\begin{subfigure}[b]{0.32\linewidth}
		\centering
		\includegraphics[height=3.2cm]{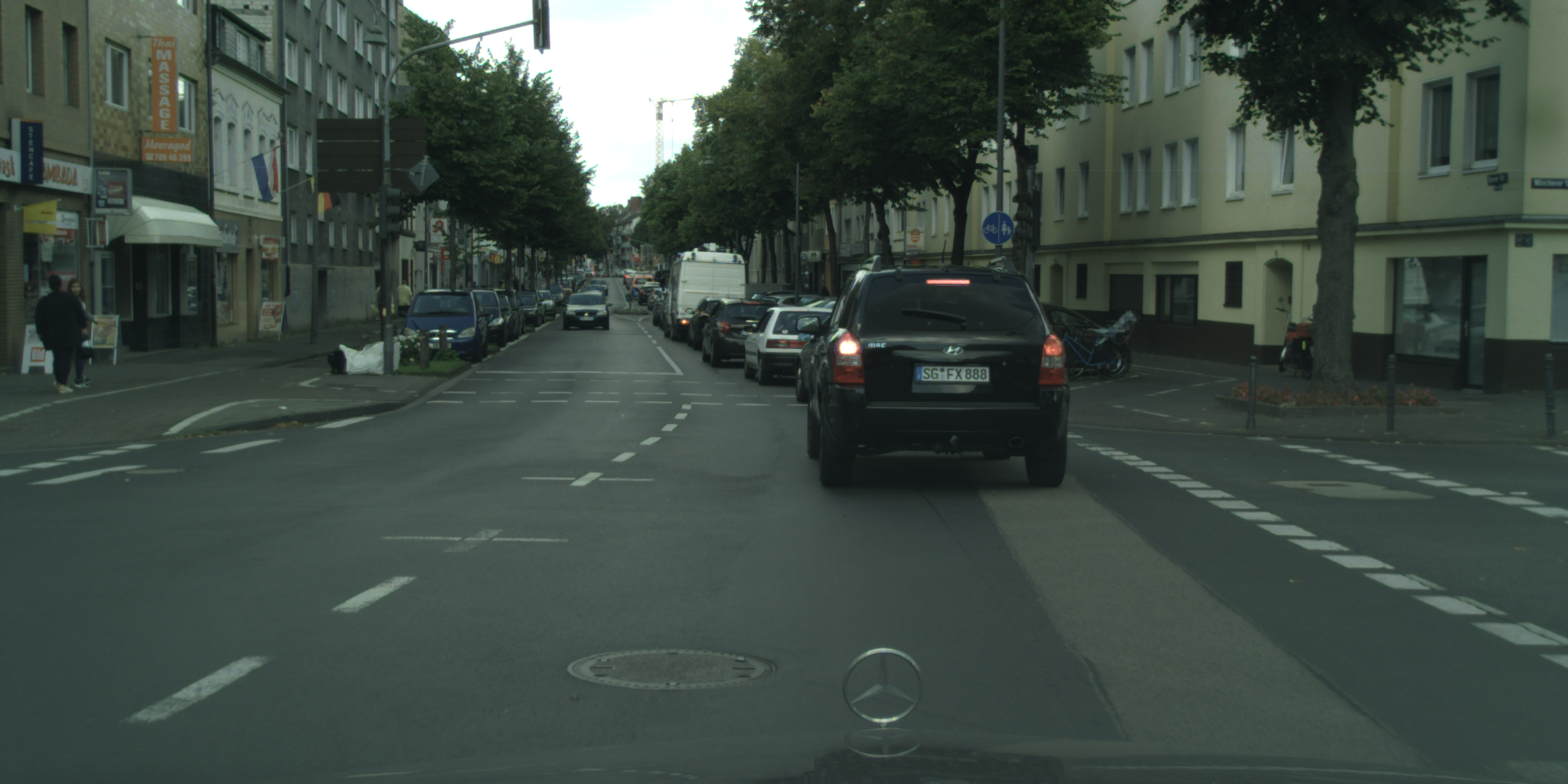}
		\caption{Cityscapes}
		\label{fig:citscapes_epe_gtav:cityscapes}
	\end{subfigure}
	\begin{subfigure}[b]{0.32\linewidth}
		\centering
		\includegraphics[height=3.2cm]{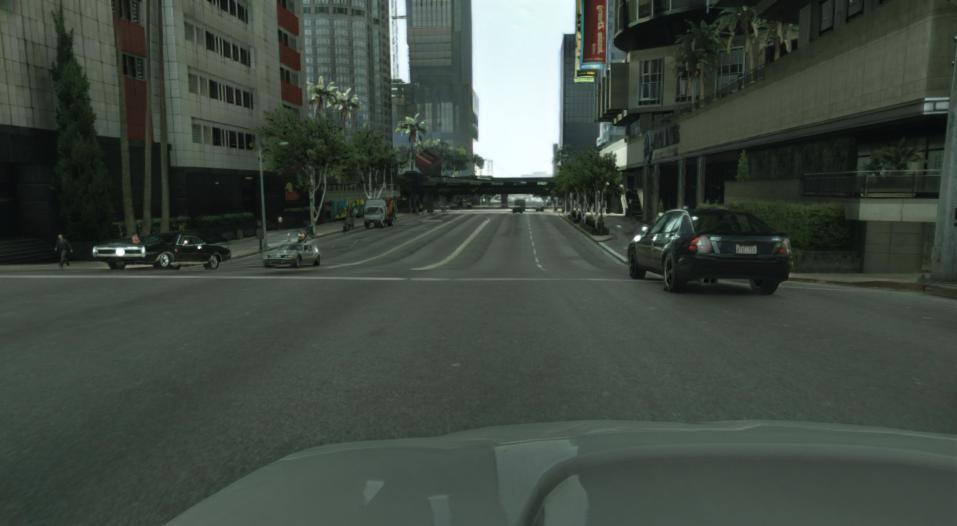}
		\caption{GTAV-EPE}
		\label{fig:citscapes_epe_gtav:epe}
	\end{subfigure}
	\begin{subfigure}[b]{0.32\linewidth}
		\centering
		\includegraphics[height=3.2cm]{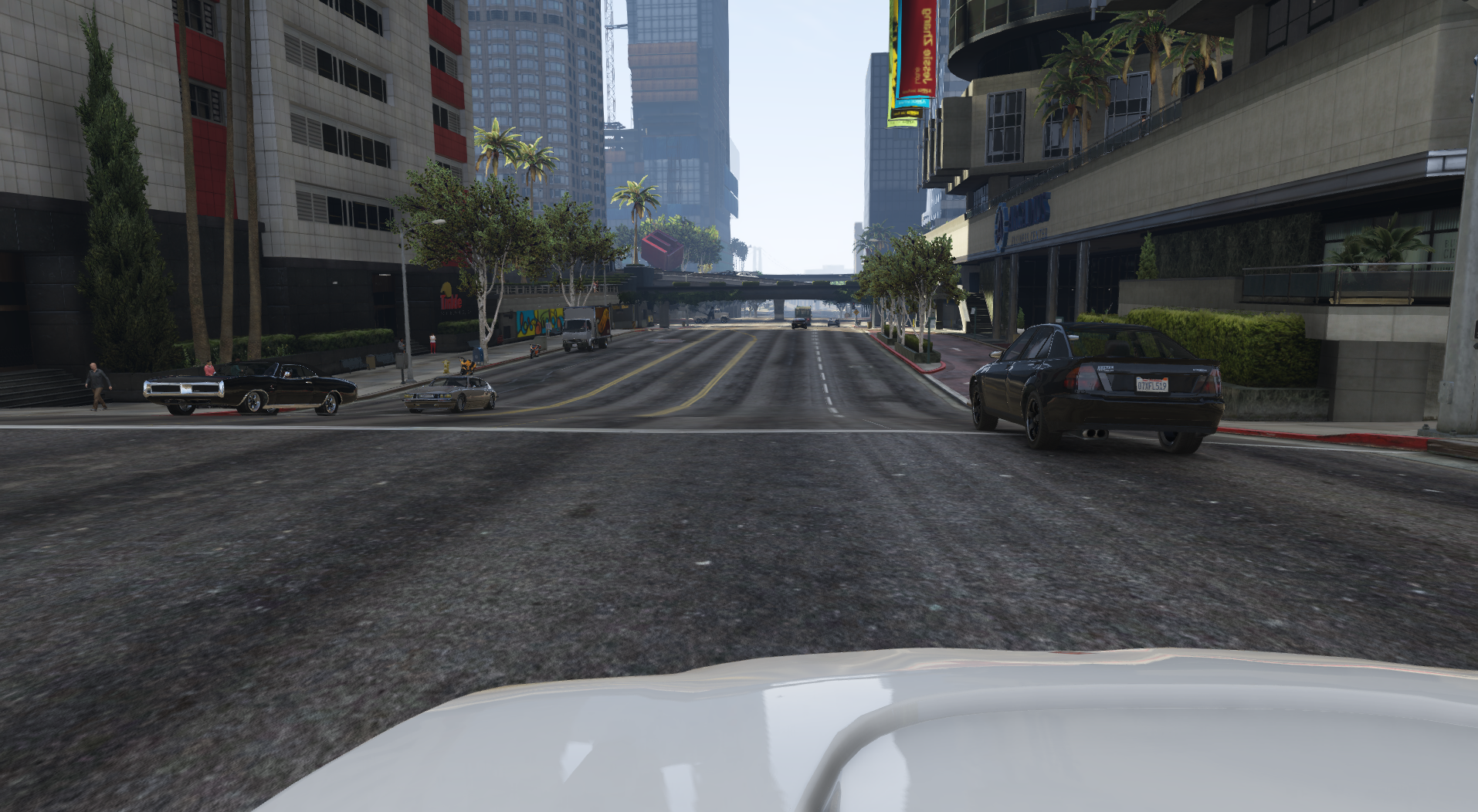}
		\caption{GTAV}
		\label{fig:citscapes_epe_gtav:gta}
	\end{subfigure}
	\caption{Datasets demonstrated by EPE \cite{richter2021enhancing}. Left: Cityscapes \cite{cordts2016cityscapes}, center: GTAV-EPE, right: GTAV \cite{richter2016playing}}
	\label{fig:citscapes_epe_gtav}
\end{figure*}

In their EPE paper, Richter et al. \cite{richter2021enhancing} demonstrated significant improvements over other GAN work, with attributes that correlate well with the interests of simulation in robotics, including the ability to minimize artifacts, retain temporal consistency, and prevent the altering of the ground truth of the image. Alongside their paper, they provided a dataset of enhanced images produced by their trained network. These image were enhanced from GTAV~\cite{richter2016playing} to be visually similar to Cityscapes~\cite{cordts2016cityscapes}. To illustrate this, we provide an image from Cityscapes (Fig.~\ref{fig:citscapes_epe_gtav:cityscapes}) next to an image from GTAV (Fig.~\ref{fig:citscapes_epe_gtav:gta}) and its EPE-enhanced pair (Fig.~\ref{fig:citscapes_epe_gtav:epe}). In this section, we evaluate the data produced in EPE by quantifying how well it predicts the real-world performance of a real-data trained network, and how well it serves as a stand-in for reality when used to train a segmentation algorithm. From here, the EPE-enhanced GTAV data generated and provided by EPE will be referred to as GTAV-EPE.

\subsection{Segmentation networks and datasets}

There is great potential in the idea of using simulation to produce synthetic segmentation data that can be used for training and testing perception algorithms for on-road driving. Therefore, we look to compare Cityscapes, GTAV, and GTAV-EPE data from the following perspective: to what extent could GTAV or GTAV-EPE data be used instead of Cityscapes data for the same application? 

Insofar as ``the application'' is concerned, first we consider a semantic segmentation task since semantic labels exist for both Cityscapes and GTAV. Since the ground truth in GTAV-EPE is unchanged from GTAV, we can recycle the semantic maps between the datasets. For perception, we chose to use NVIDIA's semantic segmentation work~\cite{tao2020semanticsegmentation} and their associated library \cite{nvidiaSemanticSegmentation} as a representative algorithm that has been demonstrated to achieve high accuracy on Cityscapes segmentation. Furthermore, the library is distributed with a pre-trained network trained on the Cityscapes dataset.

While a pre-trained network accelerates the evaluation of the synthetic data for testing an existing algorithm, we also need variants of the network that are trained on GTAV and GTAV-EPE in order to understand any potential benefit when training on the enhanced synthetic data. To this end, we trained the segmentation network on both GTAV and GTAV-EPE data. As we are interested in quantifying the similarity between simulation and reality, we do not consider using simulation to supplement real data. Supplementation using simulation is the subject of improvements to network training rather than improvements to simulation. Since we do not have knowledge of the specific augmentations and datasets used for the pre-trained network, we further retrained a Cityscapes variant of the network with the same training procedure as that for GTAV and GTAV-EPE, and with no coarse labels used from Cityscapes due to the lack of equivalence in GTAV. 
For appropriate subsequent comparisons, we split the datasets into training, validation, and test subsets to ensure no cross-contamination between any training and testing of perception algorithms. For each dataset, approximately 600 images were set aside for testing/comparison. The rest were split between training and validation, with the majority used for training. As we do not know what data was included when training the GAN that produced GTAV-EPE, splits were done arbitrarily. Each network was trained to convergence on its own validation set and their same-domain test accuracy are listed in Table~\ref{tab:segmenation_accuracies}.

\begin{table}
	\centering
	\caption{Accuracy of the segmentation networks. Pre-trained accuracy is different from that reported in \cite{tao2020semanticsegmentation}, likely due to scaling of images to the size of GTAV-EPE. IOU is the mean intersection over union for all classes.}
	\begin{tabular}{ |m{3.5cm}|c|c|c| } 
		\hline
		\textbf{Segmentation Network} 	& \textbf{IOU [\%]} & \textbf{Pixel Accuracy [\%]} \\
		\hline
		Cityscapes pre-trained 	& 82.3 & 96.5 \\ 
		\hline
		Cityscapes (retrained)	& 77.0 & 96.2 \\ 
		\hline
		GTAV 					& 75.1 & 96.1 \\ 
		\hline
		GTAV-EPE 				& 69.9 & 95.2 \\
		\hline
	\end{tabular}
\label{tab:segmenation_accuracies}
\end{table}

\subsection{Generalization of contextualized performance to semantic segmentation}

To quantify the sim-to-real gap in semantic segmentation we use a validation methodology which considers the difference in predicted and real contextualized performance - CPerf~\cite{elmquist2022performance}. Therein, object detection was used, and patches of similar content were compared, with content similarity defined as a similarity in the ground-truth object bounding boxes. The patches were centered on the objects of interest and the contextualized performance was evaluated on a per-class basis.

When applied to object detection, it is natural to center patches around the objects of interest in order to associate a specific performance with a specific image patch (object). However, for the segmentation problem, no such anchors exist to determine which candidate patches should be compared. Therefore, we propose the use of a random set of $k$ patches per image, each of size $p \times p$. These candidate patches are used to find other patches with similar content (semantic labels). All patches with similarity above a threshold $t$ are then included in the batch-wise comparison. The similarity function of choice for this comparison is based on pixel similarity:
\begin{equation}
{\mathcal{S}}_{p \times p}(c,a)
=
\frac{\Sigma_{i,j} \mathds{1} \left[{\mathcal{G}}_{i,j}(c) = {\mathcal{G}}_{i,j}(a) \right]}{p^2} \; ,
\end{equation}
where $\mathds{1}$ is the indicator function that returns 1 when the condition is true, and 0 otherwise. $\mathcal{G}_{i,j}(c)$ and $\mathcal{G}_{i,j}(a)$ are the ground truth labels at pixel $i,j$ for reference patch $c$ and sample patch $a$ respectively. $p$ is the patch size. The sums runs across the entire patch, resulting in a similarity measure equivalent to pixel accuracy. One could just as easily base similarity on IOU or any other performance measure to impose higher weight on less frequent classes.

Once patches of similar content of found, we follow CPerf \cite{elmquist2022performance} and compare, batch-wise, the performance distribution on the set of similar patches. A per-batch prediction error is measured, and the mean across all batches compared is the final contextualized performance difference. See \cite{elmquist2022performance} for the full algorithm.

\subsection{Sim-to-real results}

\begin{figure}
	\centering
	\begin{subfigure}[b]{.49\linewidth}
		\centering
		\includegraphics[width=\linewidth]{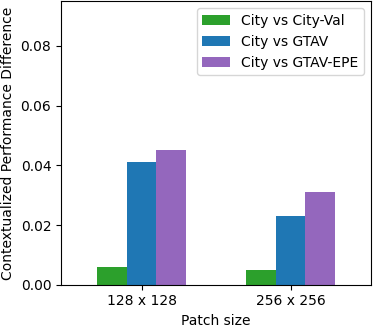}
		\caption{Testing {\SBELcitynetpt}}
		\label{fig:epe_for_testing}
	\end{subfigure}
	\begin{subfigure}[b]{.49\linewidth}
		\centering
		\includegraphics[width=\linewidth]{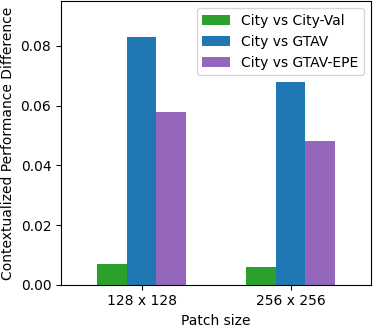}
		\caption{Testing {\SBELcitynet}}
		\label{fig:epe_for_testing2}
	\end{subfigure}
	\caption{Measure of the sim-to-real gap when evaluating the pretained (\ref{fig:epe_for_testing}) and retrained (\ref{fig:epe_for_testing2}) Cityscapes segmentation network. Smaller values are better.}
	\label{fig:epe_for_testing_all}
\end{figure}

To assess the suitability of the GTAV and GTAV-EPE data for testing, we consider the pre-trained segmentation network ({\SBELcitynetpt}) and the retrained Cityscapes network ({\SBELcitynet}). Given a real dataset (Cityscapes) and two synthetic datasets (GTAV and GTAV-EPE), we evaluate the simulations' ability to predict the real performance of {\SBELcitynetpt} and {\SBELcitynet}. Since the interest is in the sim-to-real difference, we calculate the CPerf difference between Cityscapes and GTAV, and between Cityscapes and GTAV-EPE. The comparison is performed across the images in the test sets, using patches with size $128\times128$ and $256\times256$, with 64 and 16 patches per image respectively and a similarity threshold of 0.75. The results for {\SBELcitynetpt} (Fig.~\ref{fig:epe_for_testing}) show the contextualized performance on GTAV to be more similar than that of GTAV-EPE, indicating that the enhancement does not assist in improving simulation for evaluation of this pre-trained network in this case.

While this is counter to the results seen in EPE \cite{richter2021enhancing}, a few training procedures indicated in~\cite{tao2020semanticsegmentation} provide a possible explanation, and indicate robustness to domain transfer. First, according to~\cite{tao2020semanticsegmentation}, training of {\SBELcitynetpt} used coarse labels of Cityscapes that do not precisely match the image. This likely adds robustness to the domain transfer and would limit improvements of GTAV-EPE. Second, the training process described in~\cite{tao2020semanticsegmentation} first used a secondary dataset (namely, Mapillary Vistas~\cite{neuhold2017mapillary}) to pre-train the network, adding additional robustness to domain transfer. 

With this in mind, we re-run the comparison with our own retrained Cityscapes network {\SBELcitynet} and again evaluate the contextualized performance difference between Cityscapes-GTAV and Cityscapes-GTAV-EPE. 
Figure~\ref{fig:epe_for_testing2} shows the results of using GTAV-EPE instead of GTAV when testing {\SBELcitynet}. Here we see that in fact, GTAV-EPE performs more similarly to reality than GTAV, by a factor of about 0.2 (20-30\%) when considering patches of 128x128 and 256x256 pixels. Unfortunately, there is still a gap between the performance in reality and the performance in GTAV-EPE, indicating room for improvement in the enhancement algorithm.

To understand where the performance differences are coming from when testing {\SBELcitynet}, examples of the overlapping batches are provided as supplemental material, see \cite{GANinSimSuplementalMaterial}. The GTAV examples induced segmentation errors in large parts of the road, likely due to the difference in road texture between Cityscapes and GTAV. These texture differences are mitigated by the EPE-GAN and performance in those regions becomes more similar to reality.

For assessing the ability of EPE-GAN to improve the synthetic data for training a perception network, we adopt a different methodology. We train two variants of the segmentation network, considering each dataset (GTAV and GTAV-EPE) as the synthetic data and then gauge that network's ability to transfer to the real data (Cityscapes). Critically, we are not looking for the simulation that produces the best real-performing segmentation network, but rather the simulation that produces the network with the smallest sim-to-real difference. We train the segmentation algorithm with synthetic data to produce {\SBELgtanet} and {\SBELgtaepenet}. Training followed the default scheme provided by the segmentation library.

To evaluate the sim-to-real difference of a network, we observe its performance on its own dataset vs. its performance on Cityscapes. To that end, it is important for the network to properly converge in training and to produce a test-set accuracy that is meaningful. Thereafter, we obtain two measures: (1) CPerf between Cityscapes and GTAV as observed by {\SBELgtanet}; and (2) CPerf between Cityscapes and GTAV-EPE as observed by {\SBELgtaepenet}. This comparison reveals which dataset produced a smaller sim-to-real difference. 

\begin{figure}
	\centering
	\includegraphics[width=\linewidth]{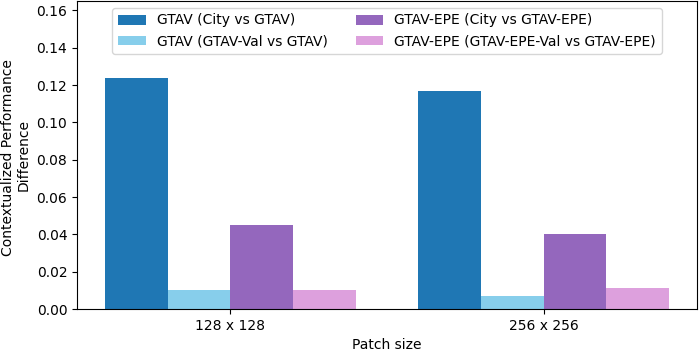}
	\caption{Quantification of the sim-to-real gap using GTAV and GTAV-EPE data for training a segmentation algorithm. Lower is better. Reference performance similarity is provided between the test and val sets to indicate intra-domain transfer.}
	\label{fig:epe_for_training}
\end{figure}

Figure~\ref{fig:epe_for_training} shows that the enhanced data produced a network with far smaller sim-to-real gap than the original GTAV data. In fact, the gap is approximately three times smaller, with a CPerf difference very similar to what {\SBELcitynet} perceived between Cityscapes and GTAV-EPE. This indicates significant potential for EPE-GAN to be used to mitigate the sim-to-real gap when training on synthetic data; however, it also shows room for improvement before the GTAV-EPE data could replace Cityscapes without reduction in quality.

The differences between the sim-to-real transfer of {\SBELgtanet} and {\SBELgtaepenet} can be seen in the supplemental material \cite{GANinSimSuplementalMaterial}. In these examples, a difference in the performance of {\SBELgtanet} was witnessed on regions of roads (often mistaken for sidewalks) and signs/poles (often entirely missed). Likely, the road errors are attributable to visually large texture differences between the datasets.

\section{GAN enhancement for simulating indoor lab}
\label{sec:epe_simulation}

This section discusses the use of the EPE-GAN for an indoor task of detecting cones, in an effort to understand how EPE-GAN can be used to enhance the simulation of the environment. As part of a larger task~\cite{artatk2022}, the desire is to both train and test a 1/6th scale autonomous vehicle in simulation and understand how it can be done such that the lessons and algorithms obtained in simulation can be transferred effectively to reality.

\subsection{Datasets and GAN training}
\label{sec:epe_simulation:training}

Simulation of the 1/6th scale autonomous vehicle by is carried out in  ProjectChrono~\cite{chronoOverview2016,projectChronoGithub}, leveraging Chrono::Vehicle~\cite{chronoVehicle2019} and Chrono::Sensor~\cite{asherSensorSimulation2021}. This enables close-loop testing of the autonomy algorithms in simulation before deploying onto the miniature autonomous vehicle. We look here at the ability of EPE-GAN, when applied to images produced by Chrono::Sensor, to produce enhanced, realistic images that provide better insights into the capabilities and deficiencies of the perception algorithms used by the vehicle. While the training and evaluation herein are performed offline, the GAN could be added to the rendering pipeline to perform inference online within Chrono::Sensor.

The baseline simulated data is generated using Chrono::Sensor, which leverages physically-based rendering \cite{burley2012physically} and real-time ray tracing \cite{optixNVIDIA} to allow modeling and simulation of cameras along with other sensors including lidar and radar \cite{asherSensorSimulation2021}. For the simulated images, the virtual environment was constructed by hand from reference images of the real environment. Included in the baseline camera simulation is a model of a wide-angle lens, based on the polynomial radial model \cite{sturm2011camera}, using parameters calibrated from the camera used on the reference vehicle \cite{artatk2022}. During data collection with the scaled vehicle, it was noted that the onboard camera did not perform white-balancing, so a color-correction model was added to the camera simulation that unbalanced the colors based on calibration images from the real camera.

The cone datasets were generated from Chrono::Sensor with images and segmentation maps. However, to use the rendering aware denormalization (RAD) modules in EPE, we must provide per-frame graphics buffers. Within the custom graphics pipeline in Chrono::Sensor, we can optionally use normal and albedo buffers to denoise stochastic path tracing results. A small modification was made to write these buffers alongside the images and semantic maps to serve as the graphics buffers in EPE. While in the EPE paper six graphics buffers were used, a conscious decision is made herein to limit the computing, memory, and complexity burden of the simulation and only provide albedo and normal buffers. Additionally, while EPE leveraged a robust segmentation map in the discriminator, this is removed in this task due the inability of existing segmentation algorithms to include the classes of interest for the application. 

\begin{figure*}
	\centering
	\begin{tabular}{M{.1\textwidth}M{.192\textwidth}M{.192\textwidth}M{.192\textwidth}M{.192\textwidth}}
		\toprule
		Source (Simulation) & 
		\includegraphics[width=.2\textwidth]{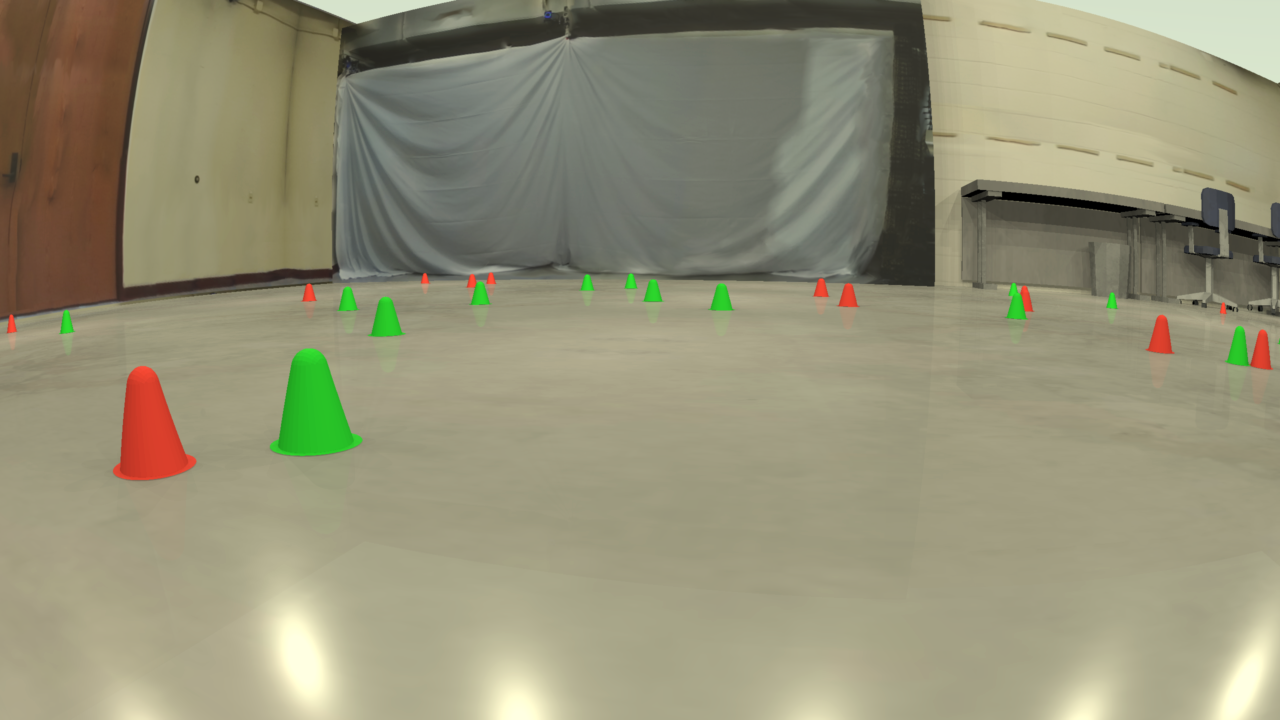} & 
		\includegraphics[width=.2\textwidth]{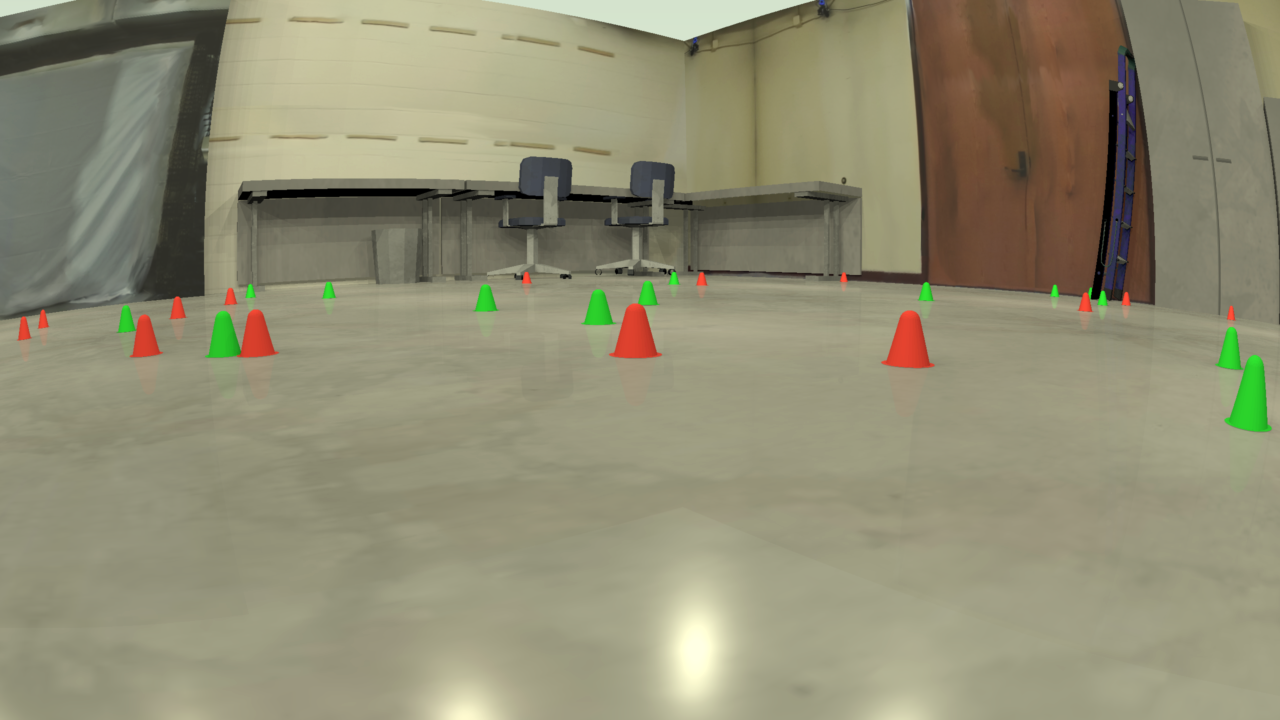} & 
		\includegraphics[width=.2\textwidth]{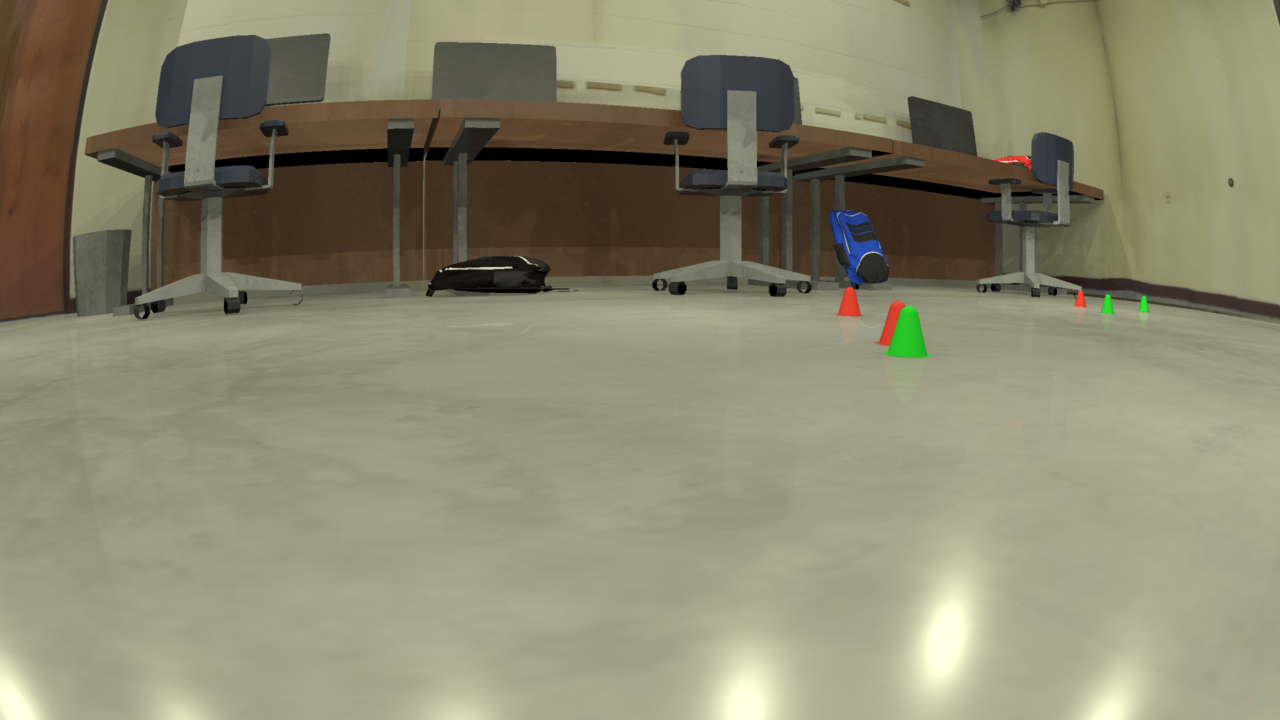} &
		\includegraphics[width=.2\textwidth]{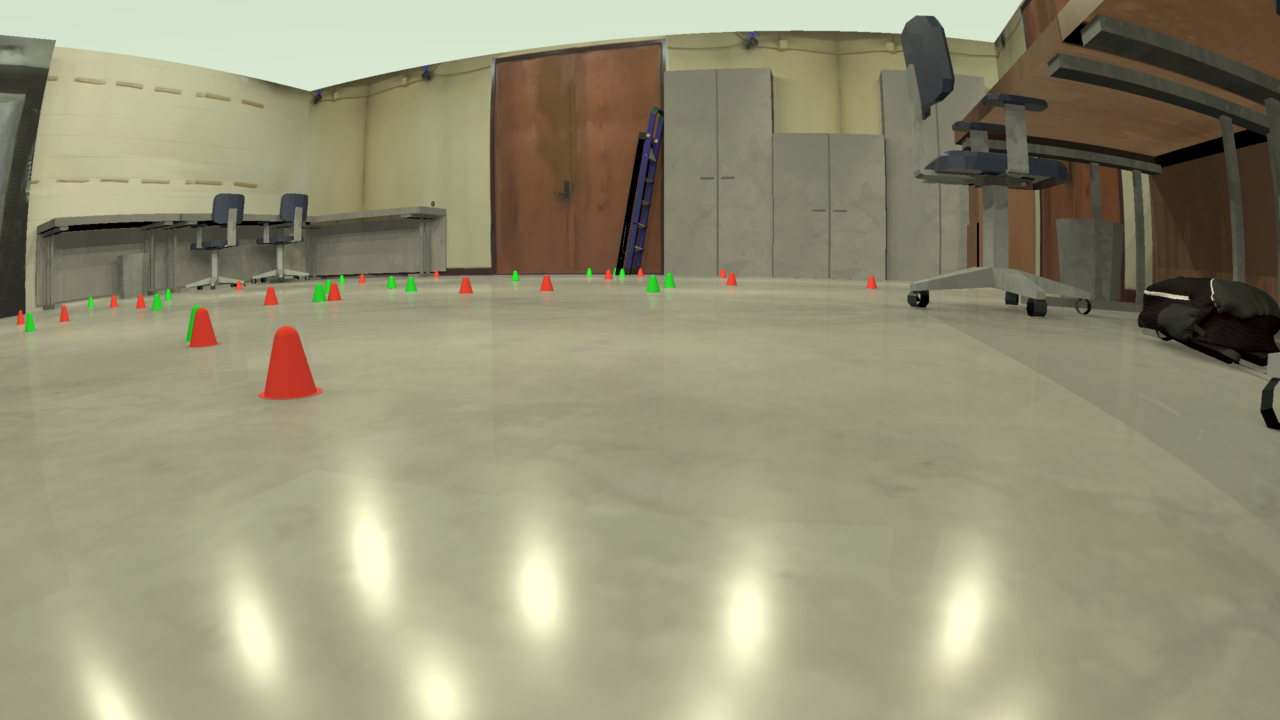} \\
		Target (Real) & 
		\includegraphics[width=.2\textwidth]{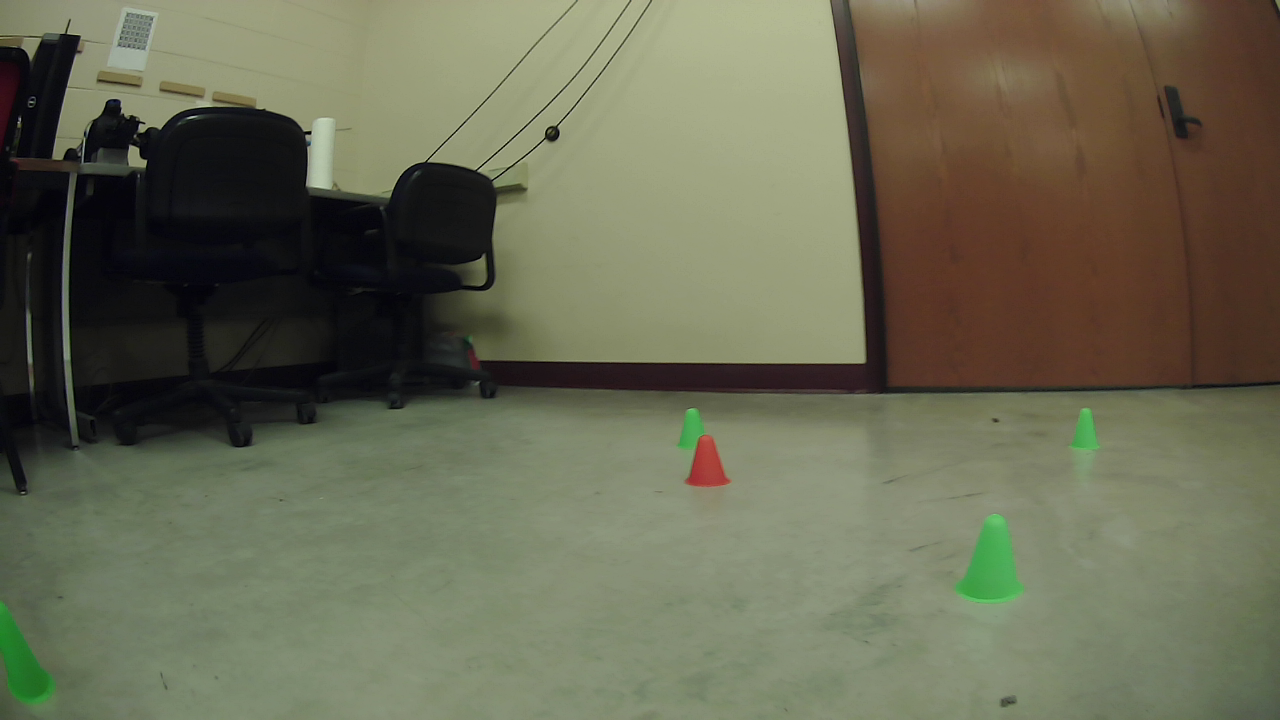}  & 
		\includegraphics[width=.2\textwidth]{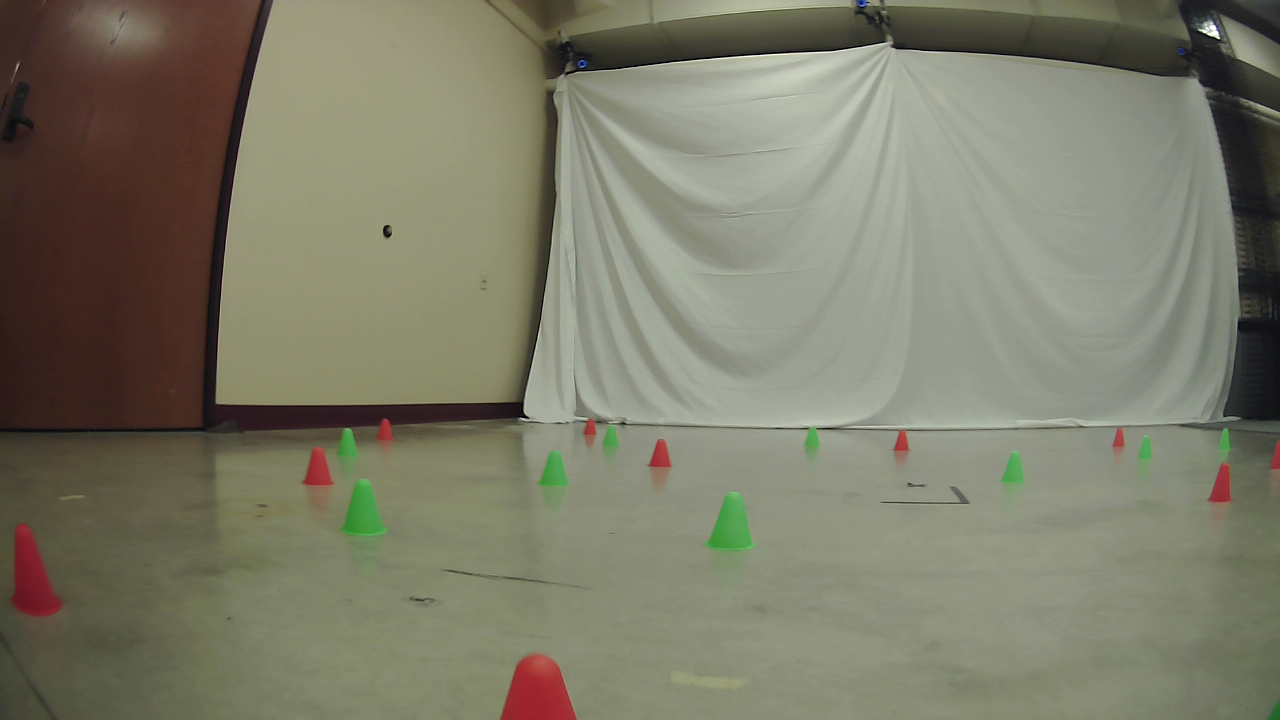} & 
		\includegraphics[width=.2\textwidth]{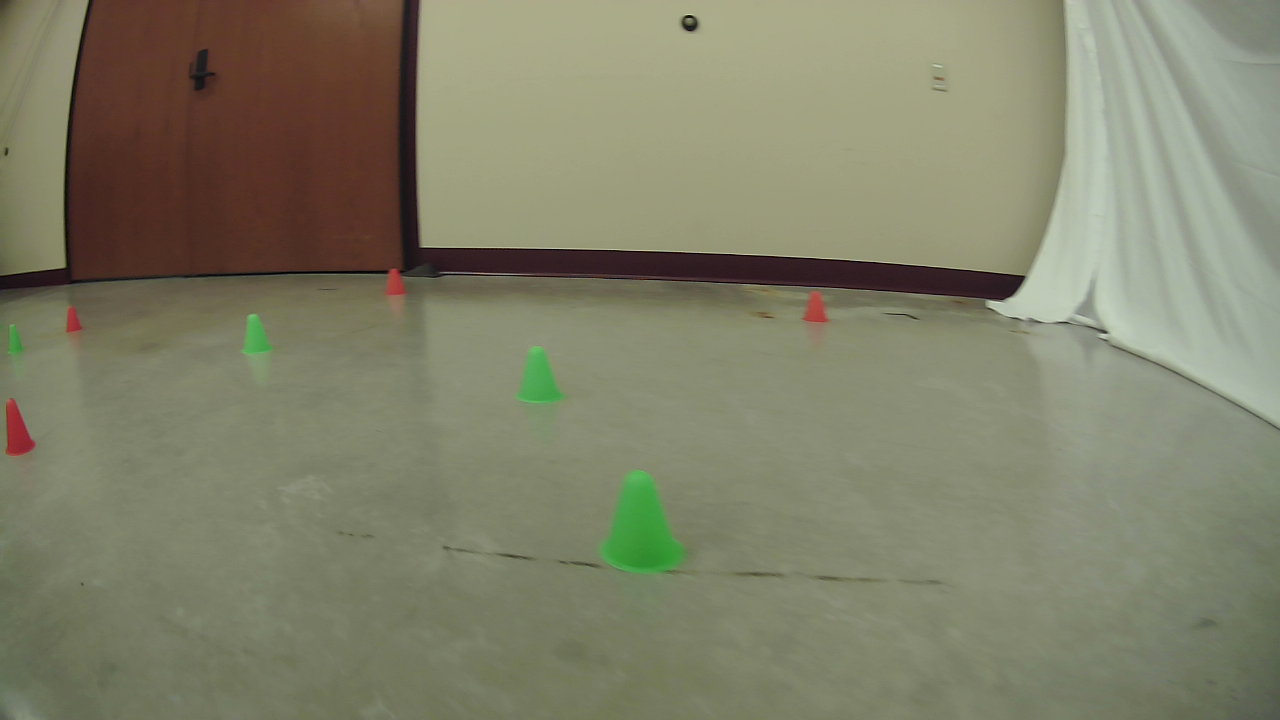} & 
		\includegraphics[width=.2\textwidth]{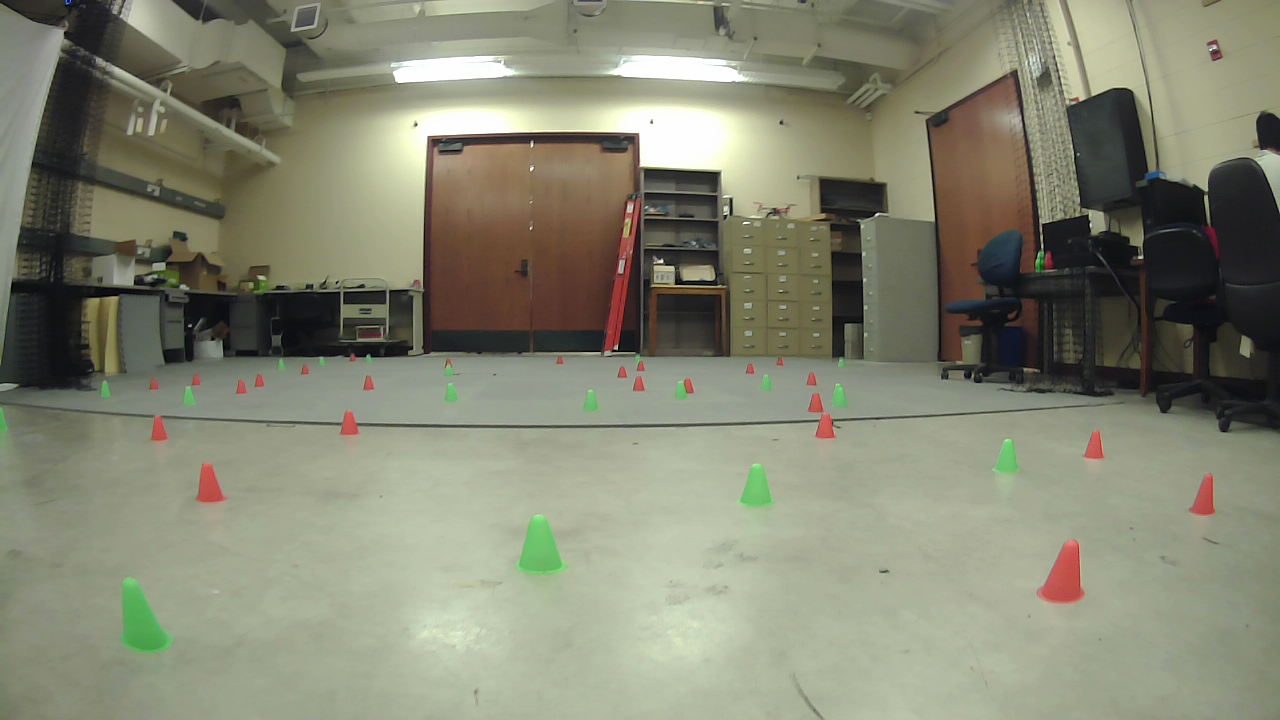}  \\
		\bottomrule
	\end{tabular}
	\caption{Example training images from the source and target datasets.}
	\label{fig:example_train_images}
\end{figure*}

Within the simulation, segmentation results are generated only for cones since these are the only regions of interest for the application. All other objects are labeled as background. Since the environment is small, the EPE-GAN training sets contained 2200 simulated (source) images and 1186 real (target) images. Each image was captured in an indoor environment with randomly distributed red and green cones; the number of green and red cones was identical. Figure~\ref{fig:example_train_images} shows examples of the source and target images. 

For training on the cone data, a modification was made to EPE to weight the matched training pairs based on the inclusion of objects of interest (cones). The weighting was set for patches containing cones to encourage a focus of the regions of interest. Variants with class weight 10 (CW10) and class weight 1 (CW1) were considered. The GAN was trained while monitoring the kernel VGG distance (KVD) (not aligned)~\cite{richter2021enhancing} on the validation set and selecting the GAN which produced the lowest KVD, while not introducing major artifacts; this resulted in 400,000 and 1,250,000 training iterations for CW10 and CW1, respectively. The learning rate was initialized to $10^{-5}$ as higher rates resulted in instability. An example image from CW10 can be seen in Fig.~\ref{fig:epe_cone_images}, with additional examples provided in the supplemental material \cite{GANinSimSuplementalMaterial}.

\subsection{Datasets and detector training}
\begin{figure*}
	\centering
	\begin{subfigure}[b]{0.32\linewidth}
		\centering
		\includegraphics[width=\linewidth]{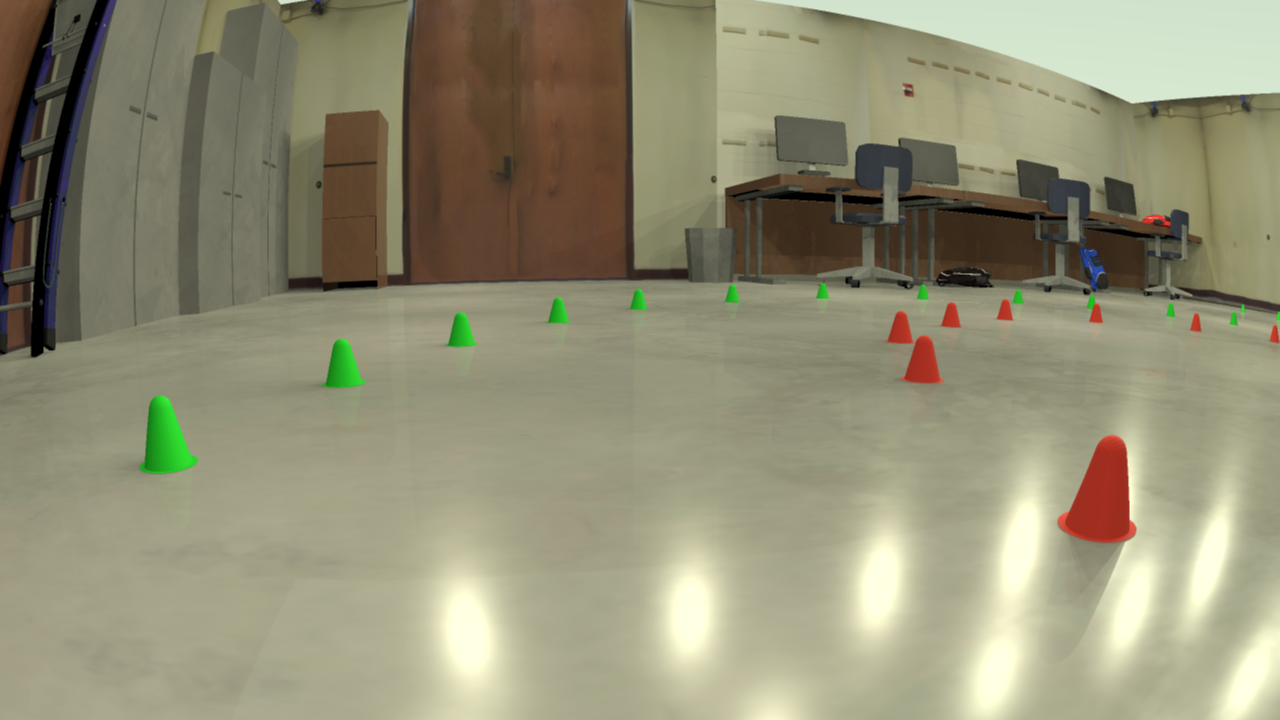}
		\caption{Simulated without augmentation.}
	\end{subfigure}
	\begin{subfigure}[b]{0.32\linewidth}
		\centering
		\includegraphics[width=\linewidth]{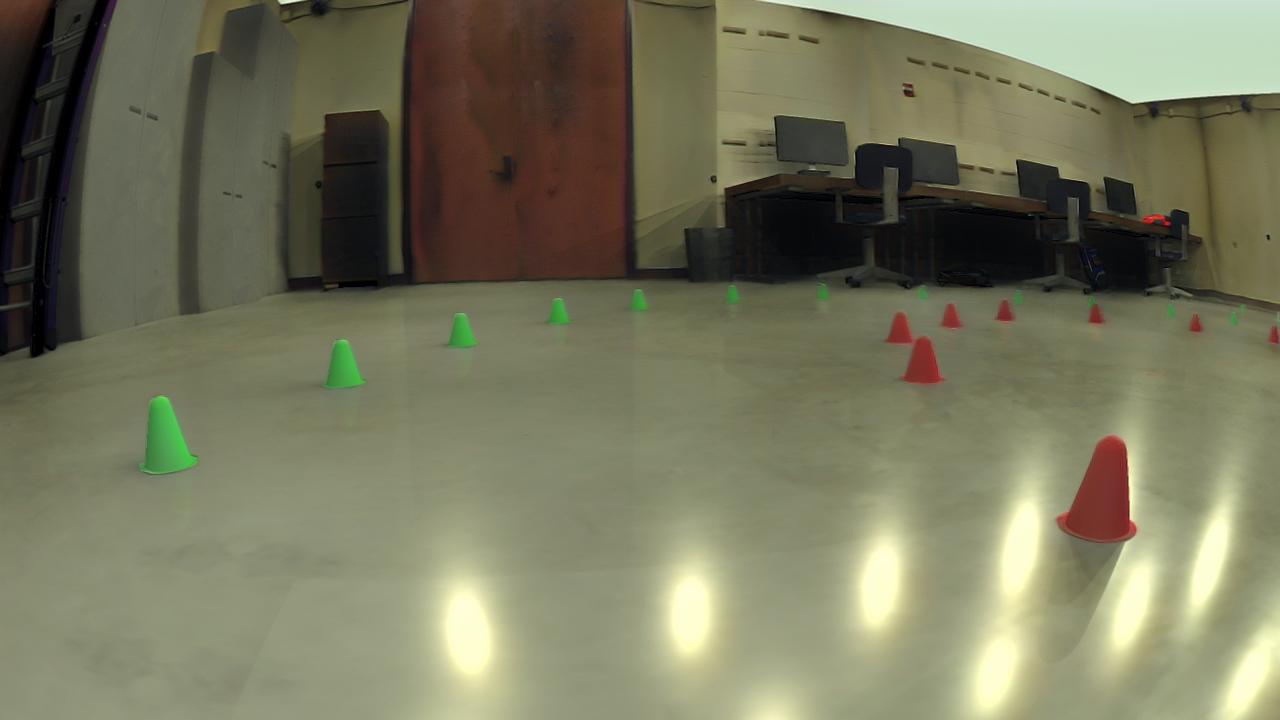}
		\caption{Simulated with EPE augmentation.}
	\end{subfigure}
	\begin{subfigure}[b]{0.32\linewidth}
		\centering
		\includegraphics[width=\linewidth]{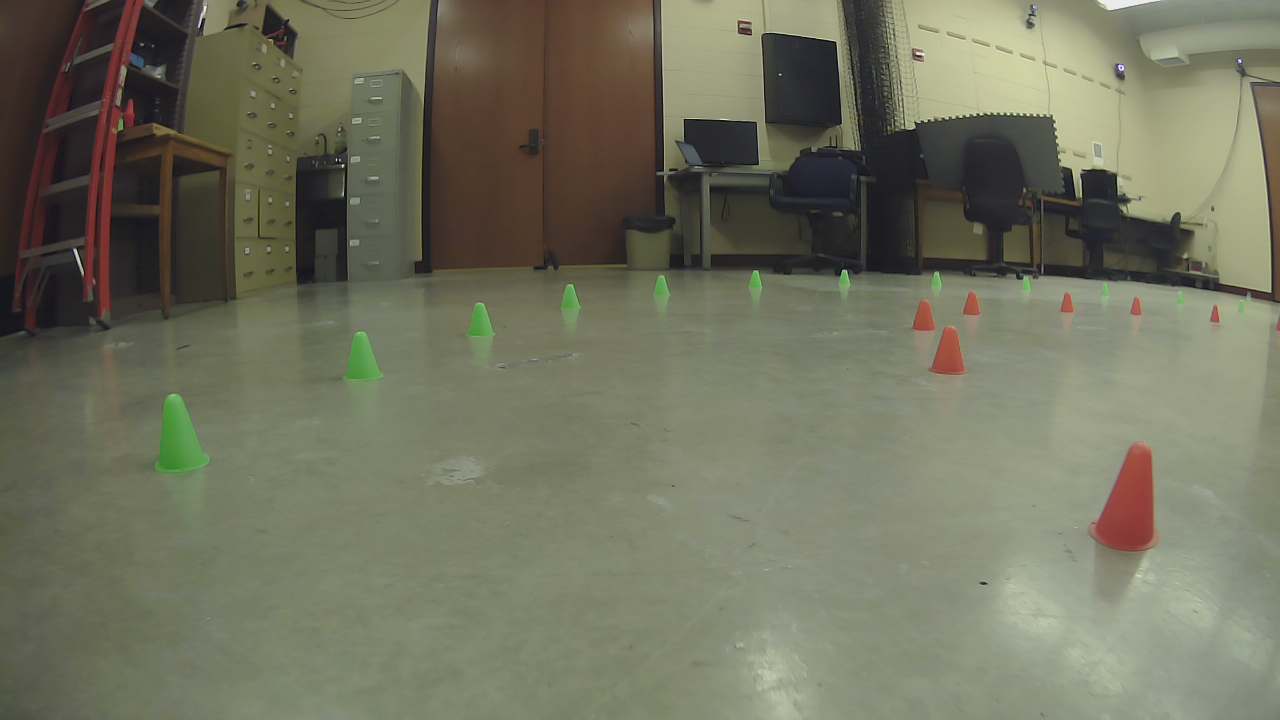}
		\caption{Real image.}
	\end{subfigure}
	\caption{Example images from metric dataset.}
	\label{fig:epe_cone_images}
\end{figure*}

The goal of the simulation is to produce synthetic data for testing and training an object detector. The detector of choice is the YOLOv5 Nano network \cite{glenn_jocher_2020_4154370} due to its accuracy and inference time on the on-vehicle processor \cite{artatk2022}. For training the object detector, a dedicated dataset was used so as not to included any data used to train the GAN. A real-world training dataset of 200 images was taken in the lab and hand labeled. Corresponding validation and test sets were obtained to determine stopping condition and final test metrics. From simulation, a training set of 500 images were generated from the lab environment with uniformly distributed cones. These images were produced alongside their graphics buffers and automatically labeled in simulation. Similar validation and test sets were obtained for the simulated case. To produce the object detector datasets for {\SBELepenet}, the simulated training, validation, and test sets were enhanced by the GAN. The metric datasets were yet again a separate dataset to ensure control. The datasets used for comparison came directly from the 1/6th scale vehicle application, where the cones defined a path for the autonomous vehicle to follow. The metric datasets, with example images shown in Fig. \ref{fig:epe_cone_images}, were reconstructed using a motion tracking setup, and were roughly paired. Misalignment was due to errors in calibration (camera intrinsic and extrinsic parameters), motion tracking, lens model, and scene reconstruction. See \cite{elmquist2022performance} for more details. The variants of the object detector were all trained with the same hyperparameters until convergence on their respective validation sets, and evaluated against their respective test sets. For each object detector, it was able to achieve, on its own domain, a mAP@.5 of 0.95-0.99 and a mAP@.95 of 0.77-0.79 (here, mAP@p represents the {\em mean average precision} at an IOU value p).

\subsection{Results and realism in training and testing}

\begin{figure}
	\centering
	\begin{subfigure}[b]{.49\linewidth}
		\centering
		\includegraphics[width=\linewidth]{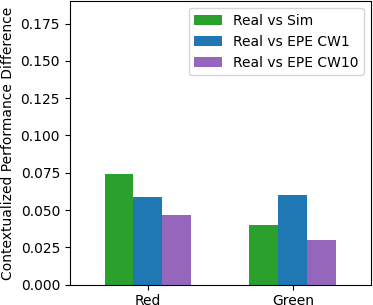}
		\caption{Testing {\SBELrealnet}}
		\label{fig:epe_for_testing_cones}
	\end{subfigure}
	\begin{subfigure}[b]{.49\linewidth}
		\centering
		\includegraphics[width=\linewidth]{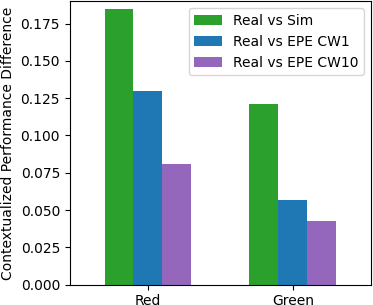}
		\caption{Training in sim}
		\label{fig:epe_for_training_cones}
	\end{subfigure}
	\caption{Left: measure of the sim-to-real gap when evaluating the real-trained network ({\SBELrealnet}). Right: measure of the sim-to-real gap when evaluating the sim or sim-epe training network ({\SBELsimnet} and {\SBELepenet}).}
	\label{fig:epe_for_testing_cones_all}
\end{figure}

With a set of trained detectors: {\SBELrealnet}, {\SBELsimnet} and {\SBELepenet}, we evaluate next their performance differences. For evaluation, we first consider the use of simulation for testing the performance of {\SBELrealnet}. To do so, we evaluate the performance difference of {\SBELrealnet} between real data and sim data, and between real data and sim-EPE data. These differences are obtained using a contextualized performance with patch twice the size of the cone, similarity threshold of 0.8, and a minimum batch size of 4. The results are shown in Fig.~\ref{fig:epe_for_testing_cones}. It is clear that {\SBELrealnet} is already fairly robust to the sim-to-real difference and that enhancing the simulation can have mixed results, given that sim-EPE CW1 data had higher difference than sim data for green cones.

For evaluating the use of sim-EPE data for training an object detector, we use the same comparison parameters, but now compare the performance difference of {\SBELsimnet} between real and sim, and the performance of {\SBELepenet} between real and sim-EPE. Figure~\ref{fig:epe_for_training_cones} shows that both variants (CW10 and CW1) improve the simulated data for training, with CW10 producing significantly better results. However, neither variant is as robust as {\SBELrealnet} to the sim-to-real differences. Examples of the batch-wise comparison are provided as supplemental material \cite{GANinSimSuplementalMaterial}. For large objects, very little difference in performance is seen, as all network variants were able to detect these with high accuracy. The major differences arose for small cones, where simulated small cones were far more distinguishable from their background than real cones. The EPE-GAN data improves these small cone representations from a performance standpoint. That is, {\SBELepenet} is not only able to better detect small real cones, but {\SBELepenet}'s behavior on its own domain's small cones is more similar to its performance on small real cones.

\section{Conclusion and future Work}
\label{sec:conclusion}

The results described here demonstrate that EPE-GAN has significant potential to improve camera simulation for robotics. We were able to demonstrate this through quantifying the perception-response of segmentation and object detection networks, considering the induced response on similar real and simulated content. The improvements seen in the results held for training of the networks for both the city and indoor environments. However, we believe that this quantitative comparison and validation is application and algorithm specific, as demonstrated when using simulation for assessment. During testing of a network that was trained using augmentations that increase robustness of the network, the benefit of EPE-GAN was significantly limited, and in some cases counter-productive.

The results suggest that future work should look to incorporate the GAN into the sensor simulation pipeline for online inference and enhancement of simulated images. We believe that machine learned methods such as this could augment physics-based sensor modeling, as domain differences not easily captured by traditional sensor simulation, such as texture variety, could be performed by the GAN, and help reduce the sim-to-real difference for simulation of robotics.

\section*{Acknowledgment}
This work was carried out in part with support from National Science Foundation project CPS1739869. Special thanks to the ARC Lab at the University of Wisconsin-Madison for their support through their motion capture facilities.

\bibliographystyle{IEEEtran}
\bibliography{BibFiles/refsAutonomousVehicles,BibFiles/refsSensors,BibFiles/refsChronoSpecific,BibFiles/refsRobotics,BibFiles/refsSBELspecific,BibFiles/refsMachineLearning,BibFiles/refsMBS}

\vfill

\end{document}